\documentclass{article}
\usepackage{spconf,amsmath,amssymb,graphicx,booktabs,url}
\usepackage[T1]{fontenc}

\usepackage{flafter}
\usepackage{multirow}

\usepackage{xcolor}
\usepackage{pifont}
\usepackage[protrusion=false]{microtype}
\usepackage[hidelinks,hyperfootnotes=false]{hyperref} 
\newcommand{\envelope}{\mbox{\footnotesize\ding{41}}}

\graphicspath{{figures/}}

\title{GroundSeek-CXR: Retrospective Evaluation-Guided Active Observation Framework for Multi-Label Chest Radiography Diagnosis}

\name{Yihang Fu$^{1,*}$, Xinyu Liu$^{2,*}$, Qing Xu$^{3,*}$, Xiaofeng Yang$^{4}$, Zhen Chen$^{5,\envelope}$}
\address{$^{1}$Yale University \quad $^{2}$A*STAR \quad $^{3}$University of Nottingham \quad $^{4}$University of Chicago \\ $^{5}$The Hong Kong Polytechnic University}

\begin{document}
\ninept
\maketitle
\thispagestyle{plain}

\begin{abstract}
Chest radiography (CXR) diagnosis is critical for thoracic disease assessment, yet challenged by the need to identify subtle lesions and co-occurring abnormalities across regions. While recent vision-language models can actively observe local regions, existing preference-based methods supervise each observation by its immediate effect on the current diagnosis, failing to capture how each observation and stopping decision contributes to the final diagnosis. To address this issue, we propose GroundSeek-CXR, an active CXR diagnosis framework that learns observation and stopping decisions by retrospectively evaluating them with the final diagnosis. Specifically, we first devise a Retrospective Diagnostic Evaluation (RDE) that scores observation and stop candidates from offline Monte Carlo tree search through independent continuations. Moreover, we design a Retrospective Gap-weighted Preference Optimization (RGPO) that learns from retrospective-score preferences weighted by their score gaps, prioritizing actions that improve diagnostic accuracy. Extensive experiments demonstrate that our GroundSeek-CXR framework outperforms state-of-the-art methods, and improves abnormal-image sample-F1 by 4.48\% over immediate-score preference learning. 
\end{abstract}


{\let\thefootnote\relax\footnotetext{$^{*}$ Equal contribution.\quad$^{\envelope}$ Corresponding author.}}

\section{Introduction}
Chest radiography (CXR) is one of the most widely used imaging examinations for thoracic disease diagnosis, and has attracted significant research attention~\cite{calli2021deep}. Accurate CXR interpretation requires detecting diverse abnormalities, ranging from diffuse findings (\textit{e.g.}, consolidation and pleural effusion) to subtle focal lesions (\textit{e.g.}, small nodules)~\cite{nguyen2022vindr}. However, the projective nature of radiography superimposes ribs, vessels and the mediastinum onto the lung fields, so subtle lesions are easily obscured and overlooked even by experienced radiologists~\cite{drew2013invisible}. Moreover, a radiograph often contains co-occurring abnormalities across regions, demanding both global assessment and careful regional inspection~\cite{kundel1975interpreting}. These challenges have given rise to the field of multi-label CXR diagnosis, which seeks to comprehensively identify co-occurring abnormalities.

\begin{figure}[!t]
  \centering
  \includegraphics[width=0.9\linewidth]{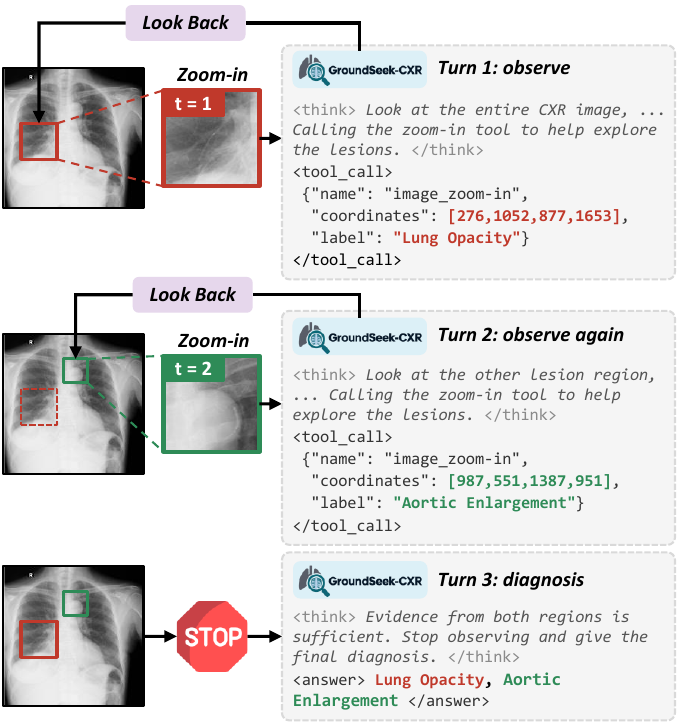}
\caption{Illustration of our GroundSeek-CXR for multi-label CXR diagnosis. At each turn, GroundSeek-CXR looks back at the radiograph and zooms into a suspicious region to observe local evidence (Turns 1-2), and stops to give the final diagnosis once the evidence is sufficient (Turn 3). Both observation and stopping decisions are learned by retrospectively evaluating them with the final diagnosis.}
  \label{fig:intro}
\end{figure}

The classical CXR diagnosis methods~\cite{rajpurkar2017chexnet,perez2025exploring} formulated this task as multi-label image classification, mapping the entire radiograph to abnormality labels in a single forward pass. More recently, medical vision-language models (VLMs)~\cite{sellergren2026medgemma,xu2025lingshu,chen2024chexagent} generated diagnoses with textual reasoning, and grounding methods~\cite{xu2025medground,mahmood2025phrase} further associated findings with anatomical regions, but they still perceived the radiograph only once at a fixed resolution. To enable regional observation, visual chain-of-thought methods~\cite{shao2024visual,hu2024visual} allowed models to crop and zoom into suspicious regions during reasoning, and ViGoRL and DeepEyes~\cite{sarch2025grounded,zheng2026deepeyes} further trained such observation behaviors with reinforcement learning. Moreover, UV-CoT~\cite{zhao2025unsupervised} learned region-conditioned reasoning through preference optimization, and ClinCoT~\cite{liu2026clincot} introduced clinical hypotheses and score-aware preferences for medical visual reasoning, while MIRA~\cite{wang2026mira} explored reflective regional inspection.

\begin{figure*}[!t]
\centering
\includegraphics[width=0.95\textwidth]{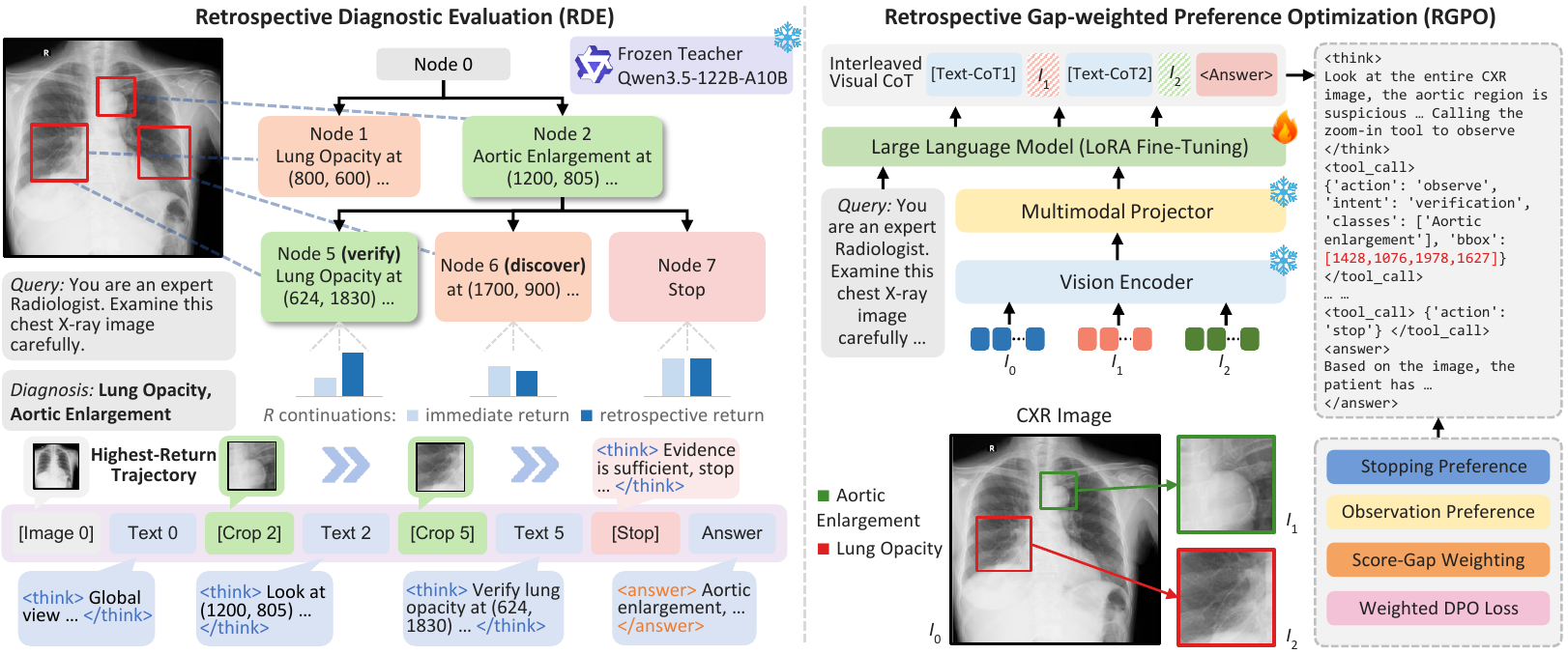}
\caption{The overview of our GroundSeek-CXR framework, consisting of the RDE and the RGPO. The RDE retrospectively scores observation and stop candidates at each search state through their independent continuations from offline MCTS. The RGPO trains the student with score-gap weighted stopping and observation preferences, enabling observation and stopping.}
\label{fig:overview}
\end{figure*}

Despite these advancements, a fundamental limitation of existing methods lies in how observations are supervised: preference-based methods~\cite{zhao2025unsupervised,liu2026clincot} evaluate each observation by its immediate effect on the current diagnosis, while reinforcement learning methods~\cite{sarch2025grounded,zheng2026deepeyes} share a single outcome reward across all decisions in a trajectory. However, in multi-label CXR diagnosis, an observation without immediate change may still guide subsequent observations toward co-occurring abnormalities. In fact, radiologists follow a hypothetico-deductive process~\cite{elstein1978medical}, iteratively inspecting regions to confirm or refute hypotheses until sufficient evidence is gathered. Therefore, both the value of each observation and the decision to stop should be judged by the final diagnosis. These insights motivate us to develop a unified framework that retrospectively evaluates both decisions through their subsequent continuations.

To overcome this bottleneck, we propose GroundSeek-CXR, an active CXR diagnosis framework that learns observation and stopping decisions by retrospectively evaluating them with the final diagnosis, as shown in Fig.~\ref{fig:overview}. Specifically, we design a Retrospective Diagnostic Evaluation (RDE) that leverages offline Monte Carlo tree search (MCTS) with a frozen teacher to propose observation and stop candidates, and evaluates each candidate by the diagnostic utility and observation cost of its independent continuations. We further introduce a Retrospective Gap-weighted Preference Optimization (RGPO) that ranks candidate actions by retrospective scores and weights each preference pair by its score gap, enabling the student to favor actions that improve the final diagnosis. Extensive experiments on VinDr-CXR and CheXpert demonstrate that the proposed GroundSeek-CXR framework outperforms state-of-the-art VLMs, and improves abnormal-image sample-F1 by 4.48\% over immediate-score preference learning with higher precision and recall.

\section{Methodology}
\subsection{Overview and Problem Formulation}
We present GroundSeek-CXR in Fig.~\ref{fig:overview}, and achieve multi-label CXR diagnosis by learning observation and stopping decisions through retrospective evaluation with the final diagnosis. To this end, we design the RDE to propose observation and stop candidates via offline MCTS with a frozen teacher and evaluate them through independent continuations, and the RGPO to train the student with score-gap weighted preference pairs, which enables the student to perform observation and stopping directly at inference without tree search.

Specifically, given a chest radiograph $I$, the student $\pi_\theta$ maintains a diagnostic state $s_t=(\hat{Y}_t,\mathcal{U}_t,\mathcal{E}_t,B_t)$ at turn $t$. Here, $\hat{Y}_t$ is the predicted label set, $\mathcal{U}_t$ contains unresolved hypotheses, and $\mathcal{E}_t$ contains acquired crops. The remaining observation budget $B_t$ is initialized to $K$. At each turn, $\pi_\theta$ selects an action $a_t\in\mathcal{A}$:
\begin{equation}
\mathcal{A}=\{\text{stop}\}\cup\{\text{observe}(p,z)\},
\label{eq:action}
\end{equation}
where $p=(x,y)$ is an integer-pixel center in the original radiograph, and $z$ describes the intended observation target in free text. Requests call \texttt{image\_zoom\_in\_tool} with \texttt{coordinates} and \texttt{label}. Centered at $p$, the tool dynamically crops a rectangle from the original-resolution radiograph. Each observation consumes one budget unit and adds the crop to $\mathcal{E}_{t+1}$. Action selection and state updates use $\pi_\theta$, conditioned on the original radiograph and acquired crops. A stop action or budget exhaustion ends the episode and returns $\hat{Y}_T$. Malformed actions or failed crops invalidate the episode.

\begin{table}[!t]
    \centering
    \setlength\tabcolsep{3pt}
    \caption{Comparison with external models on multi-label CXR diagnosis. Macro and micro precision (Pre.), recall (Rec.) and F1 (\%) are reported on VinDr-CXR, and macro metrics on CheXpert.}
    \resizebox{\columnwidth}{!}{
    \begin{tabular}{l|ccc|ccc|ccc}
    \hline
    \multirow{2}{*}{Methods} & \multicolumn{3}{c|}{VinDr (Macro)} & \multicolumn{3}{c|}{VinDr (Micro)} & \multicolumn{3}{c}{CheXpert} \\
    \cline{2-10}
     & Pre. & Rec. & F1 & Pre. & Rec. & F1 & Pre. & Rec. & F1 \\
    \hline
    Qwen3.5-4B~\cite{qwen2026qwen35} & 34.82 & 37.82 & 36.13 & 55.26 & 55.88 & 55.57 & 37.98 & 31.42 & 33.79 \\
    MedGemma-1.5~\cite{sellergren2026medgemma} & 42.72 & 46.17 & 44.31 & 61.00 & 62.42 & 61.70 & 52.03 & 45.40 & 47.94 \\
    CheXagent-2~\cite{chen2024chexagent} & 47.93 & 50.44 & 49.09 & 65.45 & 66.45 & 65.94 & 56.28 & 54.43 & 54.82 \\
    Lingshu~\cite{xu2025lingshu} & 46.27 & 49.02 & 47.55 & 63.87 & 65.02 & 64.44 & 53.45 & 51.31 & 51.80 \\
    RAD-DINO~\cite{perez2025exploring} & \textbf{62.13} & \textbf{65.38} & \textbf{63.46} & \underline{71.61} & \underline{69.61} & \underline{70.60} & \underline{61.12} & \underline{59.74} & \underline{60.02} \\
    \hline
    GroundSeek-CXR & \underline{60.21} & \underline{61.47} & \underline{59.75} & \textbf{72.71} & \textbf{73.42} & \textbf{73.06} & \textbf{65.13} & \textbf{65.46} & \textbf{64.20} \\
    \hline
    \end{tabular}}
    \label{tab:sota}
\end{table}

\begin{table}[!t]
    \centering
    \setlength\tabcolsep{3pt}
    \caption{Comparison with different training strategies on multi-label CXR diagnosis. I and T denote immediate and retrospective ranking, and U denotes uniform weights. Abnormal denotes sample-level metrics on 949 abnormal VinDr-CXR images.}
    \resizebox{\columnwidth}{!}{
    \begin{tabular}{l|ccc|ccc|ccc}
    \hline
    \multirow{2}{*}{Methods} & \multicolumn{3}{c|}{VinDr (Macro)} & \multicolumn{3}{c|}{CheXpert} & \multicolumn{3}{c}{Abnormal} \\
    \cline{2-10}
     & Pre. & Rec. & F1 & Pre. & Rec. & F1 & Pre. & Rec. & F1 \\
    \hline
    Trajectory SFT & 54.21 & 55.46 & 53.76 & 58.42 & 57.63 & 56.96 & 53.41 & 54.17 & 50.54 \\
    DPO (I/U) & 55.39 & 56.82 & 54.97 & 59.37 & 58.85 & 58.12 & 54.62 & 55.48 & 51.84 \\
    Immediate DPO (I/I) & 56.48 & 58.08 & 56.21 & 60.86 & 60.15 & 59.62 & 55.91 & 56.84 & 53.24 \\
    DPO (T/U) & \underline{58.32} & \underline{59.94} & \underline{57.95} & \underline{62.89} & \underline{62.57} & \underline{61.96} & \underline{58.26} & \underline{59.12} & \underline{55.47} \\
    \hline
    GroundSeek-CXR (T/T) & \textbf{60.21} & \textbf{61.47} & \textbf{59.75} & \textbf{65.13} & \textbf{65.46} & \textbf{64.20} & \textbf{60.41} & \textbf{60.66} & \textbf{57.72} \\
    \hline
    \end{tabular}}
    \label{tab:strategy}
\end{table}

\subsection{Retrospective Diagnostic Evaluation}
Existing methods supervise observations with either immediate effects~\cite{zhao2025unsupervised,liu2026clincot} or trajectory-level outcome rewards~\cite{sarch2025grounded,zheng2026deepeyes}, which cannot reveal how a single observation or stopping decision contributes to the final diagnosis. To address this limitation, we design the RDE that retrospectively evaluates each observation and stop candidate by the final diagnosis of its independent continuations.

Specifically, the frozen teacher runs offline MCTS on each training radiograph with upper-confidence selection, progressive widening, and mean-score backup. Each search retains up to $M$ search states, at each of which a stop candidate and at most $J$ observation candidates are proposed. Given the reference labels $Y$, the diagnostic utility of a predicted label set $\hat{Y}$ and the retrospective score of a continuation $\tau$ from turn $t$ are defined as follows:
\begin{equation}
\begin{aligned}
D(\hat{Y},Y)&=\alpha\,F_1(\hat{Y},Y)+(1-\alpha)\,\mathrm{EM}(\hat{Y},Y),\\
U_t(\tau)&=v\,D(\hat{Y}_T,Y)-\lambda_c\,N_{t:T}-\lambda_v\,(1-v),
\end{aligned}
\label{eq:return}
\end{equation}
where $\mathrm{EM}(\hat{Y},Y)$ equals 1 if $\hat{Y}$ exactly matches $Y$ and 0 otherwise, $\alpha$ balances the F1 score and the exact match, $v\in\{0,1\}$ indicates episode validity, $N_{t:T}$ counts the observation attempts from turn $t$ onward including the candidate itself, and $\lambda_c$ and $\lambda_v$ weight the observation cost and the invalidity penalty, respectively. The three terms of $U_t$ reward the final diagnosis, impose the observation cost, and penalize invalid episodes. Note that a stop candidate incurs no further observation cost, and the reference labels are only used by this offline evaluator.

Moreover, each candidate $a$ at state $s$ receives $R$ independent continuations \cite{wang2024mathshepherd} of the teacher, separate from the tree search. For the $r$-th continuation, we read out the immediate score $S^{(r)}_{\rm imm}(s,a)$ after the state update of $a$, and the retrospective score $S^{(r)}_{\rm retro}(s,a)$ after the full continuation, where both apply validity and observation cost up to the readout. The score of each candidate under horizon $h\in\{{\rm imm},{\rm retro}\}$ is averaged as:
\begin{equation}
\bar{Q}_h(s,a)=\frac{1}{R}\sum_{r=1}^{R}S_h^{(r)}(s,a).
\label{eq:qvalue}
\end{equation}
Since no further action follows a stop candidate, its immediate and retrospective scores are identical. Candidates truncated by the context or generation limit are excluded. In this way, our RDE effectively evaluates each observation and stop candidate by the final diagnosis of its continuations, providing retrospective supervision for the subsequent preference optimization.

\begin{table}[!t]
    \centering
    \setlength\tabcolsep{3pt}
    \caption{Comparison of false positives, observations per image and mean retrospective score on VinDr-CXR, including 2,051 normal and 949 abnormal images.}
    \resizebox{\columnwidth}{!}{
    \begin{tabular}{l|c|ccc|c}
    \hline
    \multirow{2}{*}{Methods} & \multirow{2}{*}{FPR (\%)$\downarrow$} & \multicolumn{3}{c|}{Obs./Image} & \multirow{2}{*}{Score$\uparrow$} \\
    \cline{3-5}
     & & Normal & Abnormal & All & \\
    \hline
    Trajectory SFT & \textbf{5.51} & 0.190 & 1.096 & 0.477 & 0.77555 \\
    DPO (I/U) & \underline{5.75} & 0.205 & 1.212 & 0.523 & 0.77703 \\
    Immediate DPO (I/I) & 6.00 & 0.210 & 1.264 & 0.543 & 0.78004 \\
    DPO (T/U) & 6.58 & 0.232 & 1.623 & 0.672 & \underline{0.78084} \\
    \hline
    GroundSeek-CXR (T/T) & 6.24 & 0.219 & 1.506 & 0.626 & \textbf{0.79134} \\
    \hline
    \end{tabular}}
    \label{tab:cost}
\end{table}

\begin{table}[!t]
    \centering
    \setlength\tabcolsep{3pt}
    \caption{Analysis of successive observations on VinDr-CXR. Abnormal-image sample-level metrics (\%) are read out at the initial state, after the first observation and at termination.}
    \resizebox{\columnwidth}{!}{
    \begin{tabular}{ll|ccc|cc}
    \hline
    \multirow{2}{*}{Methods} & \multirow{2}{*}{Readout} & \multicolumn{3}{c|}{Abnormal} & \multirow{2}{*}{FPR (\%)} & \multirow{2}{*}{Obs./Image} \\
    \cline{3-5}
     & & Pre. & Rec. & F1 & & \\
    \hline
    \multirow{3}{*}{Immediate DPO} & Initial & 53.13 & 52.84 & 49.52 & 5.36 & 0.000 \\
     & First observation & 54.78 & 55.32 & 52.04 & 5.85 & 0.320 \\
     & Final & 55.91 & 56.84 & 53.24 & 6.00 & 0.543 \\
    \hline
    \multirow{3}{*}{GroundSeek-CXR} & Initial & 53.41 & 53.18 & 49.80 & 5.41 & 0.000 \\
     & First observation & 57.48 & 58.05 & 54.84 & 6.14 & 0.333 \\
     & Final & \textbf{60.41} & \textbf{60.66} & \textbf{57.72} & 6.24 & 0.626 \\
    \hline
    \end{tabular}}
    \label{tab:readout}
\end{table}

\subsection{Retrospective Gap-weighted Preference Optimization}
Although recent methods~\cite{zhao2025unsupervised,liu2026clincot} incorporate score differences into preference optimization, their preferences are derived from the immediate quality of each response, and the decision to stop is rarely compared against further observation. To address this, we design the RGPO that ranks observation and stop candidates by their retrospective scores and weights each preference pair by its score gap.

Specifically, we first initialize the student through trajectory supervised fine-tuning (SFT) on the highest-scoring valid trajectory of each radiograph whose final label set is exactly correct, and adopt it as the frozen reference model $\pi_{\rm ref}$. For two candidates $a$ and $a'$ at the same search state $s$, including states from unsuccessful trajectories, we compute the score gap $\bar{\delta}_h=\bar{Q}_h(s,a)-\bar{Q}_h(s,a')$ for $h\in\{{\rm imm},{\rm retro}\}$. A pair is retained only if, under both horizons, $|\bar{\delta}_h|>\epsilon$ and at least $\kappa$ of the $R$ indexed score differences agree with the sign of $\bar{\delta}_h$. This shared screening ensures that immediate and retrospective supervision are compared on common candidate pairs, which cover both observation--observation and observation--stop comparisons. Since stop pairs may constitute only a small fraction $p$ of the dataset $\mathcal{D}$, we assign each pair $i$ a base weight $\eta_q(i)$ and normalize it with the retrospective score gap, as follows:
\begin{equation}
\begin{aligned}
\eta_q(i)&=
\begin{cases}
q/p, & \text{stop pair},\\
(1-q)/(1-p), & \text{otherwise},
\end{cases}\\[2pt]
w_i&=\frac{\eta_q(i)\,|\bar{\delta}_{{\rm retro},i}|}{\mathbb{E}_{j\sim\mathcal{D}}\big[\eta_q(j)\,|\bar{\delta}_{{\rm retro},j}|\big]},
\end{aligned}
\label{eq:weight}
\end{equation}
where $q$ is the target share of stop pairs, and $q=p$ recovers unit base weights. The student $\pi_\theta$ is then optimized with the weighted direct preference optimization (DPO)~\cite{rafailov2023direct} objective:
\begin{equation}
\mathcal{L}_{\rm RGPO}=-\mathbb{E}_{i\sim\mathcal{D}}\Big[w_i\log\sigma\big(\beta\,[\ell_\theta(a_i^{+}|s_i)-\ell_\theta(a_i^{-}|s_i)]\big)\Big],
\label{eq:rgpo}
\end{equation}
where $a_i^{+}$ and $a_i^{-}$ denote the preferred and dispreferred actions ranked by $\bar{\delta}_{{\rm retro},i}$, $\ell_\theta=\log\pi_\theta-\log\pi_{\rm ref}$ is computed only over action and termination tokens, and $\beta$ controls the deviation from $\pi_{\rm ref}$. Replacing the retrospective scores with immediate scores in both ranking and weighting yields the immediate-score preference learning baseline. Although only actions receive the preference loss, action selection and state update share the same decoder parameters. In this way, our RGPO effectively aligns both the direction and the strength of action preferences with the final diagnosis, enabling the student to favor observation and stopping decisions that improve multi-label CXR diagnosis under a limited observation budget.

\begin{figure}[!t]
  \centering
  \includegraphics[width=0.95\columnwidth]{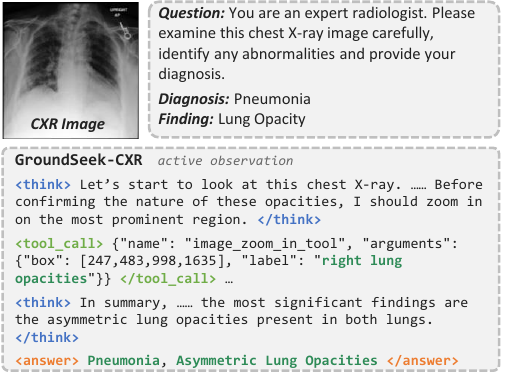}
  \caption{Visualization of our GroundSeek-CXR for multi-label CXR diagnosis. GroundSeek-CXR zooms into the suspicious right lung region to observe local evidence before stopping, and correctly identifies pneumonia with lung opacity.}
  \label{fig:case}
\end{figure}

\section{Experiments}
\subsection{Datasets and Implementations}
\noindent \textbf{Datasets}: To validate the effectiveness of the proposed GroundSeek-CXR, we conduct experiments on two benchmarks: the VinDr-CXR dataset~\cite{nguyen2022vindr}, containing 18,000 radiographs annotated with 28 abnormality labels, and the CheXpert dataset~\cite{irvin2019chexpert}, from which 202 frontal radiographs of 200 patients are used with five conditions (\textit{i.e.}, atelectasis, cardiomegaly, consolidation, edema and pleural effusion). VinDr-CXR adopts the official split, where 10\% of the training set is reserved for development and the test set contains 3,000 images (949 abnormal and 2,051 normal). The CheXpert is only used for external evaluation without training in our experiments.

\noindent \textbf{Implementation Details}: We adopt Qwen3.5-4B~\cite{qwen2026qwen35} as the student and Qwen3.5-122B-A10B as the frozen teacher on NVIDIA H200 GPUs. After adapting the vision encoder and projector on MIMIC-CXR~\cite{johnson2019mimic} and VinDr-CXR, only LoRA~\cite{hu2022lora} on the LLM decoder is trained. MCTS runs 24 iterations with $M=8$, $J=3$ and $R=4$ continuations at temperature 0.8. We set $(\alpha,\lambda_c,\lambda_v,\epsilon,\kappa,\beta)$ to $(0.75,0.02,0.20,0.01,3,0.1)$ and select $q$ on the development set. $K$ is sampled from $\{1,2,4,6\}$ in training and set to 6 at inference. RAD-DINO~\cite{perez2025exploring} adopts a VinDr-CXR-trained classifier. I, T and U denote immediate ranking, retrospective ranking and uniform weights. Results are averaged over three seeds, with abnormal-image sample-F1 as the primary endpoint and normal-image false-positive rate (FPR) for specificity.

\subsection{Comparison with State-of-the-art Methods}
To comprehensively evaluate the performance of GroundSeek-CXR, we conduct extensive comparisons with state-of-the-art methods on multi-label CXR diagnosis. As shown in Table~\ref{tab:sota}, general and medical VLMs~\cite{qwen2026qwen35,sellergren2026medgemma,chen2024chexagent,xu2025lingshu} exhibit limited performance, since they perceive the radiograph only once at a fixed resolution. RAD-DINO~\cite{perez2025exploring} with a VinDr-CXR-trained classifier achieves the highest VinDr-CXR macro-F1 of 63.46\%. In contrast, our GroundSeek-CXR outperforms all VLMs in precision, recall and F1 across three benchmark summaries, and surpasses RAD-DINO by 2.46\% micro-F1 on VinDr-CXR and 4.18\% macro-F1 on CheXpert. Notably, CheXpert is not used for training, demonstrating the generalization of the learned observation and stopping decisions. As illustrated in Fig.~\ref{fig:case}, GroundSeek-CXR zooms into the suspicious lung region before the final diagnosis, providing interpretable evidence for its predictions.

Moreover, we compare our GroundSeek-CXR with different training strategies under the same student and SFT initialization, as shown in Table~\ref{tab:strategy}. Immediate-score preference learning (Immediate DPO) evaluates each action by its immediate effect, and improves abnormal-image sample-F1 by 2.70\% over trajectory SFT. In contrast, the proposed GroundSeek-CXR framework achieves the best performance across all metrics, improving abnormal-image sample-F1 by 4.48\% over Immediate DPO with 4.50\% and 3.82\% gains in precision and recall, respectively. It further improves VinDr-CXR and CheXpert macro-F1 by 3.54\% and 4.58\%, respectively. These comparisons validate that retrospective evaluation with the final diagnosis provides more effective supervision for observation and stopping decisions than their immediate effects.

\subsection{Ablation Study}
To investigate the effectiveness of the RGPO, we conduct ablation studies on the preference ranking and gap weighting, as illustrated in Table~\ref{tab:strategy} ($2^{nd}$--$5^{th}$ rows), where all variants share the same pairs and training budget. Replacing immediate with retrospective ranking improves abnormal-image sample-F1 by 3.63\% and 4.48\% under uniform and score-gap weights, respectively, proving that retrospective scores identify more reliable preferred actions. Notably, retrospective ranking with uniform weights even outperforms Immediate DPO by 2.23\%. Moreover, score-gap weighting improves sample-F1 by 1.40\% and 2.25\% under immediate and retrospective ranking, respectively. As shown in Table~\ref{tab:cost}, under retrospective ranking, gap weighting further reduces the observations per image from 0.672 to 0.626 and the normal FPR from 6.58\% to 6.24\%. These results prove that coordinating retrospective ranking and gap weighting plays a pivotal role in our RGPO, aligning both the direction and the strength of action preferences with the final diagnosis.

\subsection{Analysis of Successive Observations}
To further analyze how observations contribute to the final diagnosis, we read out predictions at the initial state, after the first observation and at termination, as shown in Table~\ref{tab:readout}. Both policies start from similar abnormal-image sample-F1 of 49.52\% and 49.80\%. GroundSeek-CXR reaches 54.84\% after the first observation, and subsequent observations bring a further 2.88\% gain, versus 1.20\% for Immediate DPO. Moreover, as shown in Table~\ref{tab:cost}, GroundSeek-CXR allocates more observations to abnormal images (1.506 vs. 1.264 per image) while keeping similar observations on normal images (0.219 vs. 0.210), with a slight normal FPR increase from 6.00\% to 6.24\% and the highest mean retrospective score. These results demonstrate that retrospective evaluation encourages observations whose value emerges in subsequent inspection, consistent with our motivation.

\section{Conclusion}
In this work, we identify that existing preference-based active observation methods evaluate each observation by its immediate effect, and propose GroundSeek-CXR, an active CXR diagnosis framework that learns observation and stopping decisions by retrospectively evaluating them with the final diagnosis. The framework integrates the RDE that leverages offline MCTS with a frozen teacher to evaluate observation and stop candidates through independent continuations, and the RGPO that ranks candidate actions by retrospective scores and weights each preference pair by its score gap. Extensive experiments demonstrate that the proposed GroundSeek-CXR framework outperforms state-of-the-art VLMs and immediate-score preference learning with higher precision and recall, validating the effectiveness of retrospective evaluation for multi-label CXR diagnosis.

\label{endtechnical}
\clearpage

\section{Acknowledgments}
This work is supported by PolyU Start-up Fund (P0060371).

\bibliographystyle{IEEEbib}
\bibliography{references}
\end{document}